\documentclass[conference]{IEEEtran}
\usepackage{cite}
\usepackage{amsmath,amssymb,amsfonts}
\usepackage{algorithmic}
\usepackage{graphicx}
\usepackage{textcomp}
\usepackage{xcolor}

\def\BibTeX{{\rm B\kern-.05em{\sc i\kern-.025em b}\kern-.08em
    T\kern-.1667em\lower.7ex\hbox{E}\kern-.125emX}}

\usepackage{hyperref}

\usepackage{kantlipsum}

\usepackage{multirow}

\usepackage{algorithm2e}
\RestyleAlgo{ruled}

\usepackage[capitalise]{cleveref}

\begin{document}

\title{Federated Learning With Individualized Privacy Through Client Sampling}

\author{\IEEEauthorblockN{1\textsuperscript{st} Lucas Lange}
\IEEEauthorblockA{\textit{ScaDS.AI Dresden/Leipzig} \\
\textit{Leipzig University}\\
Leipzig, Germany \\
lange@informatik.uni-leipzig.de}
\and
\IEEEauthorblockN{2\textsuperscript{nd} Ole Borchardt}
\IEEEauthorblockA{\textit{ScaDS.AI Dresden/Leipzig} \\
\textit{Leipzig University}\\
Leipzig, Germany \\
ob14dupe@studserv.uni-leipzig.de}
\and
\IEEEauthorblockN{3\textsuperscript{rd} Erhard Rahm}
\IEEEauthorblockA{\textit{ScaDS.AI Dresden/Leipzig} \\
\textit{Leipzig University}\\
Leipzig, Germany \\
rahm@informatik.uni-leipzig.de}
}

\maketitle

\begin{abstract}
With growing concerns about user data collection, individualized privacy has emerged as a promising solution to balance protection and utility by accounting for diverse user privacy preferences. Instead of enforcing a uniform level of anonymization for all users, this approach allows individuals to choose privacy settings that align with their comfort levels. Building on this idea, we propose an adapted method for enabling Individualized Differential Privacy (IDP) in Federated Learning (FL) by handling clients according to their personal privacy preferences. By extending the SAMPLE algorithm from centralized settings to FL, we calculate client-specific sampling rates based on their heterogeneous privacy budgets and integrate them into a modified IDP-FedAvg algorithm. We test this method under realistic privacy distributions and multiple datasets. The experimental results demonstrate that our approach achieves clear improvements over uniform DP baselines, reducing the trade-off between privacy and utility. Compared to the alternative SCALE method in related work, which assigns differing noise scales to clients, our method performs notably better. However, challenges remain for complex tasks with non-i.i.d. data, primarily stemming from the constraints of the decentralized setting.
\end{abstract}

\begin{IEEEkeywords}
differential privacy, federated learning, privacy-preserving machine learning, user privacy, personalization
\end{IEEEkeywords}

\section{Introduction}\label{sec:intro}
    With the adoption of Machine Learning (ML) in everyday life and the increasing demand for representative training data, concerns about data security and privacy for contributing individuals have become more significant~\cite{yao2024survey,liu2021machine,yeom2018privacy,lange2024sliceupunmaskinguser}. In this context, decentralized training algorithms, such as Federated Learning (FL), which ensure that data remains on devices (e.g., smartphones), and provable privacy guarantees, such as Differential Privacy (DP), are gaining prominence~\cite{kairouz:2021,zhou2021guest,ponomareva2023dp,apple:2018,tezapsidis:2017}.
    
    FL addresses the issue of centralizing training data but does not inherently guarantee data privacy, as attacks on model parameters remain possible~\cite{boenisch2023curious}. To mitigate this, proposed modifications provide DP guarantees by introducing noise into clients' model updates~\cite{mcmahan:2018}. However, DP introduces a trade-off between privacy and model utility, where optimizing this trade-off is critical for usability~\cite{lange2024assessing}. Since individuals generate data with varying sensitivity levels and privacy requirements, Individualized Differential Privacy (IDP) has emerged as a promising approach, allowing users to select privacy levels (e.g., low, medium, or high)~\cite{berendt2005privacy}. This enables models to learn more effectively from data with lower privacy requirements, rather than applying the strictest privacy guarantees uniformly.
    
    Existing IDP algorithms in FL focus on local DP~\cite{yang:2021}, impose additional restrictions~\cite{shen:2023}, or adjust noise multipliers~\cite{aldaghri:2023}. However, such approaches have shown suboptimal results in centralized DP training~\cite{boenisch:2023}. Instead, findings suggest that individualized sampling rates, which alter the probability of including data in a training step, offer better performance.
    
    We propose an approach to leverage individualized client sampling rates in FL, enabling personalized privacy guarantees for clients. To transfer the sampling method from centralized to federated settings, we introduce an updated training algorithm that determines client participation based on their privacy requirements. Additionally, we address challenges such as the realistic distribution of training data and privacy budgets across clients in our experiments. Our results show that individual guarantees, applied through our sampling techniques improve the privacy-utility trade-off of standard DP and also outperform the alternative technique of noise multiplier scaling~\cite{aldaghri:2023}.
    
    \Cref{sec:background} provides an overview of the fundamentals, while \cref{sec:related} reviews related work. \Cref{sec:algo} introduces our algorithm, and \cref{sec:exp,sec:results} present the experimental setup and results, respectively. Finally, we discuss and summarize our findings in \cref{sec:discussion,sec:conclusion}.

\section{Background}\label{sec:background}

    \subsection{Federated Learning}
        Federated Learning (FL)~\cite{mcmahan:2016} is a decentralized ML approach that enables multiple participants, referred to as clients, to collaboratively train a shared model without transferring their local data to a central server. Instead, each client performs training locally, and only model updates, such as gradients or parameters, are shared for aggregation. The most common approach is \texttt{FedAvg}~\cite{mcmahan:2016} that averages across client gradients to update the global model. This process ensures that raw data remains on device, addressing privacy concerns and reducing the risks associated with data centralization~\cite{kairouz:2021}.
    
    \subsection{Differential Privacy}
        Differential Privacy (DP)~\cite{dwork:2006} is a mathematical framework to provide formal privacy guarantees when analyzing or sharing data. It ensures that the inclusion or exclusion of a single individual's data in a dataset has a limited impact on the output of an algorithm, thereby protecting individual privacy. A mechanism \( M \) satisfies \((\varepsilon, \delta)\)-DP if, for datasets \( D_1 \) and \( D_2 \) differing in at most one element, and for all output subsets \( S \):
        \[
        \Pr[M(D_1) \in S] \leq e^\varepsilon \cdot \Pr[M(D_2) \in S] + \delta ,
        \]
        where the privacy loss \( \varepsilon \) controls the strength of the privacy guarantee, with smaller values indicating stronger privacy. Parameter \( \delta \) accounts for the probability of privacy failing.

    \subsection{Differentially Private Stochastic Gradient Descent}
        In DP-compliant algorithms, the core idea is to introduce controlled noise into the computation to obscure the contribution of single data points at the cost of overall utility. Differentially Private Stochastic Gradient Descent (DP-SGD)~\cite{abadi:2016} is an optimizer adaptation that ensures that ML training satisfies DP. It works by first clipping individual gradients to a fixed norm, which limits their influence on the model updates. Then, random noise is added to the aggregated gradients to obscure contributions before finally updating the model parameters.

\section{Related Work}\label{sec:related}
    This section reviews the state-of-the-art through relevant previous work in the context of FL with DP and IDP.
    
    In general, IDP allows for varying privacy budgets across users, which may improve utility due to not applying the strictest setting to everyone, while still respecting user-specific privacy requirements. For centralized settings, Boenisch et al.~\cite{boenisch:2023} propose two methods, \texttt{SAMPLE}~\cite{jorgensen:2015} and \texttt{SCALE}~\cite{alaggan:2016}, where \texttt{SAMPLE} adjusts sampling probabilities for data points, while \texttt{SCALE} focuses on varying the added noise per data point. In their evaluation, they show that \texttt{SAMPLE} slightly outperforms on differing privacy budget distributions for their user group simulations, which are inspired by earlier studies on user behavior~\cite{jensen:2005,acquisti:2005}. Both studies underscore the gap between users’ stated privacy concerns and their actual behaviors, highlighting the need for accessible mechanisms.

    While FL as a decentralized learning setup inherently reduces privacy risks by avoiding direct data sharing, it remains vulnerable to attacks such as membership inference and reconstruction attacks~\cite{boenisch2023curious}.
    DP provides a robust defense against such threats and McMahan et al.~\cite{mcmahan:2018} thus extend the aggregation of client weights with a private \texttt{DP-FedAvg} variant by clipping client gradients and adding Gaussian noise.
        
    Shifting FL to IDP mainly shifts the view from data points to a client-level privacy perspective. As in central settings, IDP enables better model utility through allowing user-specific privacy requirements.
    \cite{yang:2021} and~\cite{shen:2023} propose methods for local DP with individualized budgets, which leads to noising gradients already outside of aggregation and limits potential performance. In~\cite{shen:2023} they additionally rely on limiting the $\varepsilon$-DP guarantee to a range $\tau$.
    \cite{liu:2021} focuses on reducing communication costs by projecting private client updates through gradient projection. They combine this notion with standard DP-SGD but halt training for clients once their privacy budgets are exhausted.
    Aldaghri et al.~\cite{aldaghri:2023} implement individualized noise multipliers for clients, which is evaluated at two privacy levels and translates the \texttt{SCALE} method of~\cite{boenisch:2023} to FL.

    Our method should be free of additional restrictions from local DP and thus favors a global approach with central aggregation steps for orchestration, different from~\cite{yang:2021} and~\cite{shen:2023}.
    Our closest related work in FL~\cite{aldaghri:2023}, already implemented a comparable solution to the \texttt{SCALE} method from the central setting. However, \cite{boenisch:2023} found that \texttt{SAMPLE} performed on par or better in their central evaluation.
    We thus elevate the \texttt{SAMPLE} approach from the central to the FL scenario, and thereby from data point-level privacy to client-level privacy. For this we revise the \texttt{DP-FedAvg} algorithm to an \texttt{IDP-FedAvg} variant. We further extend the evaluation from~\cite{aldaghri:2023} through more realistic privacy distributions from user studies~\cite{jensen:2005,acquisti:2005}.

\section{Client Sampling Algorithm}\label{sec:algo}

    In this section, we develop the \texttt{IDP-FedAvg} algorithm to train ML models with heterogeneous privacy guarantees in FL. The algorithm implements IDP through customized sampling rates within \texttt{DP-FedAvg}, inspired by the non-FL methods in~\cite{jorgensen:2015,boenisch:2023}. The main algorithm consists of two steps:
    
    \begin{enumerate}
        \item \textbf{Privacy Step:} Compute the noise multiplier and individualized sampling rates for each client, as described in \cref{alg:boenisch-sample}. Sampling rates are derived from client-specific privacy budgets using the \texttt{SAMPLE} algorithm by~\cite{boenisch:2023}. We adapt this procedure by mapping epochs to FL rounds and training samples to clients, transitioning guarantees from data point-level to client-level privacy.
        \item \textbf{Training Step:} In \cref{alg:idp-fedavg}, training is performed using \texttt{DP-FedAvg} aggregation with adaptive clipping~\cite{andrew:2021}. However, clients are sampled based on their individualized sampling rates before a global noise multiplier is applied uniformly across all sampled clients.
    \end{enumerate}

    In~\cite{boenisch:2023}, the authors demonstrate that training with individualized sampling rates derived from a user's specified privacy budget may achieve better results than individual scaling of noise multipliers or clipping norms. Their algorithm for the central training scenario samples data points with differing probabilities while maintaining the expected value per epoch and applying a constant noise multiplier. 
    
    This implementation can be directly adapted to calculate sampling rates for clients, as shown in \cref{alg:boenisch-sample}. Clients are grouped into $P$ groups according to their respective privacy budgets, where $|P|$ is the number of unique values. The function \texttt{GetGroupSamplingRates} outputs a uniform noise multiplier $\sigma_{\text{SAMPLE}}$ and individualized sampling rates $\{q_1, \dots, q_P\}$, which describe the probability of selecting each client for a training round. To determine these values, the algorithm starts with an initial $\sigma_{\text{SAMPLE}}$ and calculates the corresponding intermediate $\{q_1, \dots, q_P\}$ needed to satisfy each client's $\varepsilon$ using this multiplier. 
    As a condition, the expected sampling rate $q$ is based on the number of clients participating in training each round. The initial multiplier $\sigma$ corresponds to the strictest privacy budget $\varepsilon_1$. If the resulting average sampling rate across clients does not satisfy $q$, the noise multiplier is too large for the given privacy budget distribution. The noise multiplier is then iteratively reduced using a scaling factor $s_i \lessdot 1$ (slightly smaller) until the resulting rates comply. \texttt{getSampleRate} originally uses a modified PyTorch Opacus function that ensures that privacy budgets are exhausted after $I$ iterations of sampling. We translate that to the dp-accounting library for our Tensorflow version. \cite{boenisch:2023} prove that their sampling algorithm satisfies $(\{\varepsilon_1, \dots, \varepsilon_P\}, \delta)$-DP, extending their DP guarantees to groups of data points with heterogeneous sampling rates at constant noise.
    
    The training in \cref{alg:idp-fedavg} uses a modified version of standard \texttt{DP-FedAvg} with adaptive clipping by~\cite{andrew:2021}. But instead of uniformly sampling the client subset $S$ for each training round, the calculated sampling rate for each client from \cref{alg:boenisch-sample} is used. Each client thereby acquires only the noise needed for their respective individual privacy guarantee. We focus on global aggregation and do not include hyperparameters e.g. for client training in this representation.

    \SetKwComment{Comment}{// }{}
    \begin{algorithm}[tb]
    	\caption{\texttt{GetGroupSamplingRates}: calculating per-group sampling rates regarding privacy budgets, as presented in Algorithm 2~\cite[p.6]{boenisch:2023} but adapted to clients instead of data points. Subroutine \texttt{getSampleRate} translates DP noise multipliers to sampling rates as in Algorithm 4~\cite[p.17]{boenisch:2023}.}
    	\label{alg:boenisch-sample}

    	\KwIn{Per-group target privacy budgets $\{\varepsilon_1, \dots, \varepsilon_P\}$, target $\delta$, iterations $I$, sampled clients per round $c$, total number of clients $N$, per-privacy group clients $\{|\mathcal{G}_1|, \dots, |\mathcal{G}_P|\}$}
    	\KwOut{Sampling noise multiplier $\sigma_\text{SAMPLE}$, sampling rates $\{q_1, \dots, q_P\}$}
    	
    	\SetKwProg{Fn}{Function}{:}{}
    	\SetKwFunction{getSampleRate}{getSampleRate}
        \SetKwFunction{getNoise}{getNoise}

        \BlankLine
            $q \leftarrow \frac{c}{N}$\;
        	$\sigma_\text{SAMPLE} \leftarrow $\getNoise($\varepsilon_1, \delta, q, I$)\;
        	
            \ForEach{$p \in [P]$}{
        	   $q_p \leftarrow$ \getSampleRate{$\varepsilon_p, \delta, \sigma_\text{SAMPLE}, I$}\;
        	  }
        	\While{$q \not\approx \frac{1}{N} \sum_{p=1}^{P} |\mathcal{G}_p| q_p$}{
        		scaling factor $s_i \lessdot 1$: $\sigma_\text{SAMPLE} \leftarrow s_i \sigma_\text{SAMPLE}$\;
                \ForEach{$p \in [P]$}{
                    $q_p \leftarrow$ \getSampleRate{$\varepsilon_p, \delta, \sigma_\text{SAMPLE}, I$}\;
                }
        	}
        	\KwRet $\sigma_\text{SAMPLE}, \{q_1, \dots, q_P\}$\;
    \end{algorithm}

    \begin{algorithm}[tb]
        \caption{\texttt{IDP-FedAvg}: implements \texttt{FedAvg} with individualized DP, incorporating client sampling rates into \texttt{DP-FedAvg} with adaptive clipping~\cite{andrew:2021}.}
        \label{alg:idp-fedavg}
    
        \SetKwFunction{LocalUpdate}{LocalUpdate}
        \SetKwFunction{ClipUpdate}{ClipUpdate}
        \SetKwFunction{AggregateUpdates}{AggregateUpdates}
        \SetKwFunction{AddNoise}{AddNoise}
        \SetKwFunction{UpdateGlobalModel}{UpdateGlobalModel}
        \SetKwFunction{AdjustClippingNorm}{AdjustClippingNorm}
    	\SetKwFunction{GetGroupSamplingRates}{GetGroupSamplingRates}
    	\SetKwFunction{Unique}{Unique}
    	\SetKwProg{Fn}{Function}{:}{}
    
        \KwIn{Per-client target privacy budgets $\{\varepsilon_1, ..., \varepsilon_N\}$, target $\delta$, rounds $I_{Fed}$, clients $\mathcal{C}$ with $|\mathcal{C}|=N$, sampled clients per round $c$, client learning rate $\eta_{\mathcal{C}}$, server learning rate $\eta_s$, clipping quantile $\gamma$, clipping learning rate $\eta_C$, global model $\theta$}
        \KwOut{Updated model $\theta'$, clipping quantile $\gamma'$}
    
        \BlankLine
        Groups: $\mathcal{E}_\mathcal{G}$ $\leftarrow$ \Unique{$\{\varepsilon_1, ..., \varepsilon_N\}$}\; 
        $\sigma_{\text{SAMPLE}}$, $Q_\mathcal{G}$ $\leftarrow$ \GetGroupSamplingRates{$\mathcal{E}_\mathcal{G}$, $\delta$, $c$, $|\mathcal{C}|$, group sizes of $\mathcal{E}_\mathcal{G}$}\;
        $\{q_1, ..., q_N\}$ $\leftarrow$ for each client $i \in \mathcal{C}$: get its sampling rate $q_i$ regarding their privacy group from $Q_\mathcal{G}$\;
        
        \ForEach{round of training in $I_{Fed}$}{
            Sample client subset $\mathcal{S}$ according to $\{q_1, ..., q_N\}$\;
            \ForEach{client $i \in \mathcal{S}$}{
                Local: $\Delta_i$ $\leftarrow$ \LocalUpdate{$i$, $\eta_{\mathcal{C}}$}\;
                Norm of update: $\|\Delta_i\|$\;
                \If{$\|\Delta_i\| > \gamma$}{
                    Clip: $\Delta_i$ $\leftarrow$ \ClipUpdate{$\Delta_i$, $\gamma$}\;
                }
            }
            Aggregate: $\Delta$ $\leftarrow$ \AggregateUpdates{$\mathcal{S}$}\;
            Noise: $\Delta$ $\leftarrow$ \AddNoise{$\Delta$, $\sigma_{\text{SAMPLE}}$}\;
            Global: $\theta'$ $\leftarrow$ \UpdateGlobalModel{$\theta$, $\Delta$, $\eta_s$}\; 
            $\gamma'$ $\leftarrow$ \AdjustClippingNorm{$\gamma$, $\eta_C$}\;
        }
        \Return{$\theta'$, $\gamma'$}\;
    \end{algorithm}

\section{Experimental Setup}\label{sec:exp}
In this section, we detail how we conduct our experiments.

    \subsection{Implementation Environment}
        Reference code is available from our repository at \url{https://github.com/luckyos-code/flidp}. We use Python 3.10 with Tensorflow for creating ML models. The Tensorflow Federated library supports our simulations regarding FL, while our DP implementations use Tensorflow Privacy. For hardware we run our experiments on a cluster using an NVIDIA RTX 2080 Ti GPU, 64GB of memory, and an AMD EPYC 7551P CPU.

    \subsection{Federated Datasets}
        We test on common benchmarking datasets from related work: FMNIST~\cite{caldas2018leaf}, CIFAR-10~\cite{krizhevsky:2009}, and SVHN~\cite{netzer:2011}.
        Since CIFAR-10 and SVHN are focused on central scenarios, we have to transform them into FL-conform counterparts that are split across clients. For this, we also have to consider creating i.i.d. and non-i.i.d. versions to address both paradigms. In FL, i.i.d. refers to samples that are independent (do not influence another) and identically distributed at clients. In contrast, non-i.i.d. data lacks these properties, with dependencies between samples and varying distributions, as in real-world scenarios.

        The Federated MNIST (FMNIST) dataset contains about 380.000 grayscale images of handwritten digits and is an non-i.i.d. Extended MNIST version, where the digits are grouped by their respective authors across 3,383 clients. For an i.i.d. setup, we randomly distribute samples across clients.
        
        CIFAR-10 focuses on 60,000 color images divided into 10 classes of vehicles and animals. To adapt this dataset to FL, we follow the CIFAR-100 sibling, which has an non-i.i.d. variant. Thus, we sample data for each of the 500 clients using the Pachinko Allocation Method~\cite{li2006pachinko}, which unevenly distributes classes. For i.i.d., we use the same method as for FMNIST.

        \begin{figure}[tb]
            \centerline{\includegraphics[width=\linewidth]{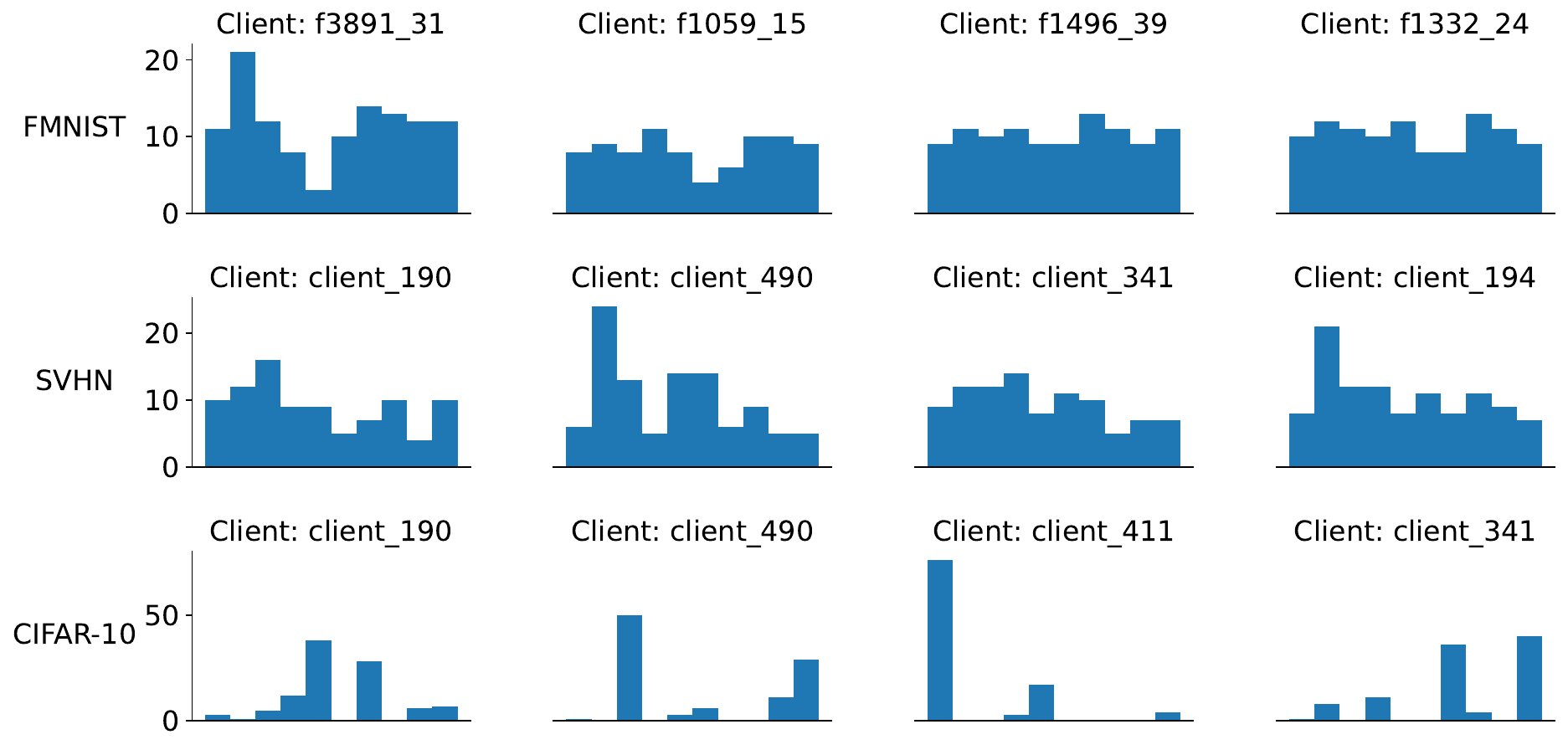}}
            \caption{Examples of label distribution on clients for our datasets (non-i.i.d.).}
            \label{fig:label_dis}
        \end{figure}
        
        The SVHN dataset features over 600,000 color images of house numbers extracted from Google Street View. However, there is no non-i.i.d. federated version and we therefore only assume our random i.i.d. distribution over 725 clients. But as shown in \cref{fig:label_dis}, we still see some skew due to some labels being overrepresented in SVHN. We can also expect a very difficult task from CIFAR-10 due to its uneven distribution. On a final note, we use the original test sets for each dataset.

    \subsection{Training Parameters} 

        \begin{figure}[tb]
            \centerline{\includegraphics[width=\linewidth]{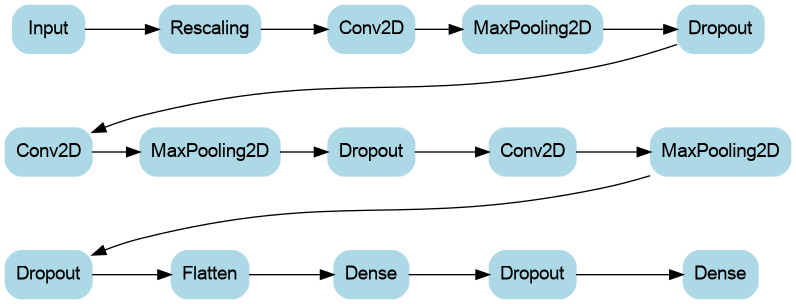}}
            \caption{CNN model architecture used in our experiments.}
            \label{fig:model}
        \end{figure}

        Our ML model is a simple Convolutional Neural Network (CNN) with 14 layers as presented in \cref{fig:model}.
        For FL server hyperparameters, we generally train for 420 rounds, sample 30 clients each round, and use a server learning rate of 1.0 for all datasets. At the clients, we use a batch size of 128 and learning rate of 0.0005 over 15 epochs. For adaptive clipping we use the standard parameter setup from~\cite{andrew:2021}.


    \subsection{Privacy Distributions}
        Regarding the evaluated privacy distributions, we incorporate real-world IDP, as well as, standard DP and non-DP settings.
        For simulating these distributions, we create three privacy groups of differing $\varepsilon$-values like $\varepsilon: \textit{1-2-3}$ and assign them across clients accordingly.
        As in~\cite{boenisch:2023}, we also rely on existing user studies for our realistic privacy distributions, which are \textit{34\%-43\%-23\%}~\cite{acquisti:2005} and \textit{54\%-37\%-9\%}~\cite{jensen:2005}.
        We further also include the strictest \textit{100\%-0\%-0\%} and most relaxed \textit{0\%-0\%-100\%} DP settings using only the respectively ranked $\varepsilon$ from the privacy groups.
        To put these differentially-private results into context we also train federated models without privacy restrictions \textit{0\%-0\%-0\%} (non-private) as our last distribution.

\section{Results}\label{sec:results}

    \begin{table}[tb]
    	\centering
        \caption{Accuracies (\%) of the model setups on all datasets}
    	\begin{tabular}{|l|c|c|c|c|c|c|}
    		\hline
    		\multirow{3}{6em}{$\varepsilon$-distribution} & \multicolumn{2}{c|}{FMNIST} & \multicolumn{2}{c|}{SVHN} & \multicolumn{2}{c|}{CIFAR-10} \\
             & \multicolumn{2}{c|}{$\varepsilon: \textit{1-2-3}$} & \multicolumn{2}{c|}{$\varepsilon: \textit{10-20-30}$} & \multicolumn{2}{c|}{$\varepsilon: \textit{10-20-30}$} \\
    		\cline{2-7}
    		& i.i.d. & non & i.i.d. & non & i.i.d. & non \\
    		\hline
    		\textit{0\%-0\%-0\%} & 98.2 & 97.0 & 86.6 & -- & 61.0 & 44.2 \\
    		\textit{0\%-0\%-100\%} & 96.5 & 95.1 & 80.9 & -- & 50.1 & 30.1 \\
    		\textit{34\%-43\%-23\%} & 95.9 & 94.6 & 79.3 & -- & 44.6 & 26.5 \\
    		\textit{54\%-37\%-9\%} & 95.4 & 93.5 & 78.2 & -- & 41.9 & 25.4 \\
    		\textit{100\%-0\%-0\%} & 93.9 & 90.7 & 76.5 & -- & 38.4 & 16.8 \\
    		\hline
    	\end{tabular}
    	\label{tab:all-results}
    \end{table}

    We gather our experimental results in \cref{tab:all-results}, where we give the percentage accuracy values for the respective models regarding the different privacy distributions. We test on all three datasets with their i.i.d. and non-i.i.d. versions. With our results, we confirm the effectiveness of our adapted algorithm. This can be derived from the stepwise change in accuracies between the privacy distributions, where \textit{0\%-0\%-0\%} poses the least strict (non-private) and \textit{100\%-0\%-0\%} the strictest. While the strictest DP level gives the lowest performance outcomes, our individualized methods put the realistic distributions in between and closer to the relaxed DP assumption of the weakest $\varepsilon$-group. In non-individualized real-world scenarios we would have to choose the strictest level at all times but with individual privacy groups through our client sampling method, we see an average advantage of 2.2\% for our i.i.d. and 5.7\% for our non-i.i.d. datasets, when assuming a \textit{54\%-37\%-9\%} distribution. With the slightly less strict \textit{34\%-43\%-23\%}, we see further average improvement by 1.5\% and 1.1\%.
    
    We however also see how differently DP effects our datasets. Even with loosened privacy requirements of $\varepsilon: \textit{10-20-30}$ for SVHN and CIFAR-10, there is a significant performance hit already when going from the non-private models to the most relaxed DP models in CIFAR-10. At $\varepsilon: \textit{1-2-3}$, private models failed for both datasets, resolving to random guessing accuracies. FMNIST on the other hand, shows to be less impacted overall. This is comparable to other existing work, where CIFAR-10 clearly showed to be a more difficult FL task that might ask for a higher privacy budget~\cite{sun:2021}. Additionally DP shows to have an even greater trade-off in FL than in central settings due to the smaller client datasets and resulting larger impact of noise on the global model~\cite{wei2020federated}. With CIFAR-10 being substantially smaller than the others, we can confirm this phenomena in our results. The non-private models are still able to achieve good performance levels but especially the non-i.i.d. version poses a challenge.

    Regarding related work, we can compare our results for non-i.i.d. FMNIST to~\cite{aldaghri:2023}, who used the \texttt{SCALE} variant of IDP and only two privacy groups: a non-private and a private group ($\varepsilon=0.6$). Their non-private group constitutes only 5\% of their clients and we therefore create another run to match their setup: $\varepsilon: \textit{0.6-$\infty$-$\infty$}$ at \textit{95\%-0\%-5\%}. They achieved 86.9\% under these conditions, while we climbed up to 90.8\% using the \texttt{SAMPLE} strategy. This supports that the advantage from the central setting carries over to FL by a larger margin.

\section{Discussion}\label{sec:discussion}
    Our experimental results demonstrate that individualized client sampling within FL can bridge the gap between strict privacy requirements and model utility. By successfully leveraging client-level sampling rates derived from privacy budgets (\texttt{SAMPLE}), our adapted \texttt{IDP-FedAvg} algorithm achieves significant accuracy improvements over uniform \texttt{DP-FedAvg} setups under realistic privacy distributions. We further confirm the advantage of individualizing sampling rates over noise scales in FL through our 3.9\% lead on FMNIST compared to related work in~\cite{aldaghri:2023}.
    However, our results also highlight the limitations of incorporating DP in FL for smaller and more complex datasets like CIFAR-10. These datasets suffer from higher sensitivity to the noise introduced by DP. 
    
    As others, we do not consider limitations for this approach regarding some common practical issues in FL, such as device heterogeneity and client availability or dropout during training in our experiments. For a more holistic and realistic view, removing our optimistic assumption would make reliable sampling of clients harder for the central server.

    The proposed method has several practical implications. Allowing users to define their privacy levels addresses a key concern in privacy-preserving systems: trust. By enabling personalized privacy budgets, users retain control over their data while contributing at a level that aligns with their comfort. This flexibility incentivizes participation, as users are not forced to adhere to uniform and potentially insufficiently strict privacy guarantees. Instead, they can balance their privacy preferences with their willingness to support collaborative learning efforts. From the perspective of ML practitioners, IDP in contrast to traditional approaches enables to benefit from higher-quality updates of users with more relaxed privacy settings. A key challenge for this aspect of IDP is the increased complexity for the user, which necessitates effective and fair communication of privacy risks to enable users to make informed assessments.

\section{Conclusion}\label{sec:conclusion}

    We present an adaptation of the IDP \texttt{SAMPLE} algorithm for FL, introducing individualized client sampling to enable heterogeneous privacy guarantees. Our \texttt{IDP-FedAvg} approach demonstrates significant utility improvements under realistic privacy distributions compared to traditional \texttt{DP-FedAvg}. As in central settings, the sampling method is able to outperform noise parameter scaling. However, the results also highlight the challenges of applying DP in federated settings, particularly for non-i.i.d. datasets and complex tasks. Future work could explore incorporating IDP through sampling rates into alternative aggregation mechanisms beyond \texttt{FedAvg}, which may better handle non-i.i.d. and noisy data distributions.

\section*{Acknowledgment}

    The authors acknowledge the financial support by the Federal Ministry of Education and Research of Germany and by the Sächsische Staatsministerium für Wissenschaft, Kultur und Tourismus for ScaDS.AI.
    Computations were done (in part) using resources of the Leipzig University Computing Centre.

\bibliography{paper.bib} 

\end{document}